\pdfoutput=1

\documentclass[11pt]{article}

\usepackage{EACL2023}
\newcommand{\name}{{\sc DepBERT}}
\newcommand{\dataset}{{\sc CausalGPT}}
\usepackage{times}
\usepackage{latexsym}

\usepackage[T1]{fontenc}

\usepackage[utf8]{inputenc}

\usepackage{graphicx}
\usepackage{tabularx}
\usepackage{subfloat}
\usepackage{caption}
\usepackage{subfig}
\usepackage{amsmath}

\usepackage{microtype}

\usepackage{inconsolata}

\title{Extracting Cause-Effect Pairs from a Sentence with a Dependency-Aware Transformer Model}

\author{
Md. Ahsanul Kabir$^{1}$ \quad S.M. Abrar Jahin$^{1}$ \quad Mohammad Al Hasan$^{2}$ \\
$^{1}$Indiana University - Purdue University Indianapolis \\
$^{2}$Indiana University Indianapolis \\
\texttt{\{mdkabir, ajahin\}@iu.edu}, \quad \texttt{alhasan@iu.edu}
}

\begin{document}
\maketitle
\begin{abstract}
Extracting cause and effect phrases from a sentence is an important NLP task, with numerous applications in various 
domains, including legal, medical, education, and scientific research. There are many unsupervised and supervised methods 
proposed for solving this task. Among these, unsupervised methods utilize various linguistic tools, including syntactic patterns, 
dependency tree, dependency relations, etc. among different sentential units for extracting the cause and effect phrases. 
On the other hand, the contemporary supervised methods use various deep learning based mask language models equipped with a 
token classification layer for extracting cause and effect phrases. Linguistic tools, specifically, dependency tree,
which organizes a sentence into different semantic units have been shown to be very effective for extracting semantic pairs 
from a sentence, but existing supervised methods do not have any provision for utilizing such tools within their model 
framework. In this work, we propose {\sc DepBERT}, which extends a transformer-based model by incorporating dependency tree 
of a sentence within the model framework. Extensive experiments over three datasets show that {\sc DepBERT} is better than 
various state-of-the art supervised causality extraction methods.
\end{abstract}
\section{Introduction}

Automatic extraction of cause and effect phrases from natural language text is an important task 
with enormous applications in various fields. In medical field, causal sentences are used for 
providing the cause and the effects associated with diseases, treatments, and side effects. 
Say, the sentence, ``Vitamin D deficiency contributes to both the initial insulin resistance and 
the subsequent onset of diabetes'', reflects disease causality; effectively extracting many such 
cause (deficiency of Vitamin D) and  effect (insulin resistance, diabetes) pairs~\cite{David:2002:causality, Tal:2000:Disease-Causality} from medical text 
can advance medical research. In fact, in medical research, causality analysis provides the
foundation for generating complex hypotheses on which new research can be designed. 
Causal sentences are also used in legal fields for determining
liability and responsibility. Consider a sentence like, ``The company's failure to adhere to 
safety regulations resulted in a workplace accident." Here, the causal link between the company's 
actions (not adhering to safety regulations) and the accident is evident. Methodologies for automatic extraction of such causal relationships from legal
texts can be useful for building an AI-based legal assistant. In the service field, an AI-based
chatbot can provide diagnostic services to customers if cause-effect phrases from instruction manuals can be mined effectively and accurately.

Due to the importance of causality extraction, several
works~\cite{atkinson2008discovering,lee2017disease,zhao2018causaltriad,an2019extracting, sorgente2013automatic, patterncausality, xiao:2019:graph} have been proposed, which either identify causal relations from sentences or extract cause and effect pairs to build causal networks. These existing works either use a set of known 
syntactic patterns to extract the cause and effect entities or use traditional supervised learning models (such as, SVM) to identify those entity pairs. Considering the cause and effect pairs as name entities, existing 
methods focus on entity extraction, and they performed well when the causes and effects are name entities, or noun phrases, such as, the name of diseases, medications, or genes. There is another group of research focusing on deep learning based methods for causality extraction\cite{bi-lstm-crf-2019, dasgupta-etal-original}. They mainly transform the causality extraction into a token-classification task. Within this group, Chansai etc al.~\cite{chansai2021automatic} fine tunes different learning methods~\cite{bert:Embedding,fasttext_Embedding,touvron2023llama} to make those amenable for the token classification task.
However, the major bottleneck for all these supervised models is the lack of mechanisms for incorporating linguistic tools,
such as, dependency tree, syntactic patterns, etc. in those models.

Dependency tree is an important linguistic tool encompassing grammatical structure, syntax, semantics, POS tags, and tag-to-tag interactions. It plays a pivotal role in the extraction of cause-and-effect relationships from textual data. Recent research studies~\cite{kabir_Asper, patterncausality} highlight the critical significance of these linguistic components in facilitating precise semantic relation extraction. Dependency parsers, such as SpaCy~\cite{Spacy2} and Stanza~\cite{qi2020stanza}, provide a powerful framework for dissecting the intricate connections between words and phrases within a sentence. Through syntactic analysis these parsers ease identification of causal verbs, subjects, and their corresponding objects. Additionally, POS tags and tag-to-tag interactions offer valuable contextual information that aids in disambiguating causal relationships, thereby enhancing the accuracy and reliability of extracted information. The integration of these dependency-based approaches into supervised causality 
extraction models would improve the extraction of cause and effect phrases, as we show in this paper. 

In this paper we propose a transformer based supervised method, \name, which seamlessly integrates the dependency structure of sentences into the bidirectional encoding representation of the transformer model. Our method effectively merges word-word co-occurrence, sentence semantics, and the syntactic dependency structure within the domain of the transformer's self-attention mechanism. Through this integration, \name\  consistently outperforms existing baseline methods, providing clear evidence that the incorporation of dependency structures significantly enhances the foundational building blocks of the transformer architecture for enhanced language understanding.

We claim the following two specific contributions:
\begin{itemize}
    
    \item We introduce \name, a transformer model that is sensitive to dependencies. It concurrently learns from dependency relation graphs, parts-of-speech tag sequences, and token-token co-occurrences for token classification tasks, outperforming conventional transformer-based language models in terms of performance.
    \item We develop a dataset, referred to as \dataset\ comprising 22,273 instances  that include cause terms, effect terms, and the sentences containing both terms. The primary objective of this dataset is to alleviate the scarcity of annotated datasets in the field of causality extraction.
\end{itemize}

\section{Related Works}

The tasks of extracting causal relations can generally be classified into three main categories: unsupervised~\cite{information.retrieval.1998,information.retrieval.2001}, supervised~\cite{dasgupta-etal-original,bi-lstm-crf-2019}, and hybrid approaches~\cite{question-answering-2006,Antonio:2013:baseline,xiao:2019:graph}. Unsupervised methods primarily rely on pattern-based approaches, employing causative verbs, causal links, and relations between words or phrases to extract cause-effect pairs. Supervised approaches, on the other hand, require a labeled training dataset containing pairs of cause and effect phrases, allowing the training of supervised learning models for the extraction of causal relationships between phrases.

Do et al. \cite{do2011minimally} introduced a minimally supervised approach based on a constrained conditional model framework, incorporating discourse connectives into their objective function. Dasgupta et al. \cite{dasgupta-etal-original} utilized word embeddings and selected linguistic features to construct entity representations, serving as input for a bidirectional Long-Short Term Memory (LSTM) model to predict causal entity pairs. Nguyen et al. \cite{nguyen2015relation} harnessed pre-trained word embeddings to train a convolutional Neural Network (CNN) for classifying given causal pairs. In contrast, Peng et al. \cite{peng2017cross} presented a model-based approach that leverages deep learning architectures to classify relations between pairs of drugs and mutations, as well as triplets involving drugs, genes, and mutations with N-ary relations across multiple sentences extracted from the PubMed corpus.

Despite the various supervised methods available, existing causality extraction techniques often fall short in incorporating dependency relations within deep transformer architectures. While some approaches do consider dependency relations~\cite{ahmad-etal-2021-syntax, song2022hierarchical, sachan2020syntax} to enhance language understanding and introduce external sequential knowledge into deep learning models~\cite{Wang2021}, this work represents a novel fusion of learning from sequential knowledge, dependency relations, and co-occurring tokens.

\section{Methodology}\label{definition} 

In this section, we first provide a formal discussion of token classification framework for solving the cause-effect pairs extraction. Then we provide the motivation and an overall framework of \name, our proposed model. Finally, we discuss \name's architecture in details.


\subsection{Problem formulation} 
Given, a sentence $S$ and two phrases $u$ (cause) and $w$ (effect) in $S$, such that they exhibit a causality relation within the sentential context. Let $S$ contains $N$ number of tokens which are $s_1$, $s_2$ ... $s_N$. Say, the cause phrase $u$ consists of the token $s_i ... s_K$, and effect phrase $w$ consists of the tokens $s_j ... s_L$, and there is no overlap between these two sequences of tokens. We also consider special tokens, such as, start token, end token and the padding tokens. Then we label all
the tokens in $S$ based on the following:

\[
    l_t= 
\begin{cases}
    1,& \text{if } \text{$s_t$ is a {\bf Special} token}\\
    2,& \text{if } \text{$s_t$ is a {\bf Cause} token}\\
    3,& \text{if } \text{$s_t$ is an {\bf Effect} token}\\
    4 & \text{otherwise}
\end{cases}
\]

we transform $l_i$ into a one-hot encoding vector of size $K$, denoted as $c_i$, using the one-hot-encoding method~\cite{harris2010digital,brownlee2017one}
Suppose there is a model, $\theta$, which predicts $p_1$, $p_2$ ... $p_M$, where $p_i$ represents the probability values predicted by the model for $l_i$. The token classification task objective of $\theta$ 
is to minimize the following multinomial cross-entropy loss function denoted by $\mathcal{L}$

\[\mathcal{L} = -\frac{1}{M}\sum_{i=1}^{M}\sum_{j=1}^{K}c_{i,j} \log p_{i,j}\]

\subsection{\name: Motivation and Design Justification}
Syntactic patterns are important to extract semantic relation~\cite{kabir_Asper, patterncausality} due to the fact that dependency relation plays an important role to identify semantic pairs. For instance, let there be a pattern \textit{u precipitates w}, which can be applied to extract two semantic pairs $u$ and $w$ from a sentence where $u$ and $w$ exhibit a cause effect semantic relation. However, the dependency relation and parts of speech tag for patterns can be crucial. For instance, here $u$ needs to be a subject for the verb $causes$ and $u$ needs to be a \textit{NOUN}, $causes$ a \textit{VERB}, and $w$ be another \textit{NOUN}. So for this particular pattern, dependency relation as well as \textit{NOUN, VERB, NOUN} sequence can be important as well.

Traditional BERT~\cite{bert:Embedding} lacks the capability to consider syntax and dependency relations when learning token embeddings, a shortcoming addressed by \name. This innovative approach incorporates dependency relations and POS tag sequences into a transformer model designed to be acutely aware of these linguistic dependencies. While several dependency parsers are available for converting sentences into dependency trees, some previous research~\cite{kabir_Asper}) has indicated that  the performance of syntactic dependency pattern extraction does not
depend much on the specific format of dependency tree. In this work, we have used Spacy dependency parser Spacy\cite{Spacy2} for converting a sentence to a dependency tree, as the API of this library was convenient. 


\begin{figure}[h!]
    \centering
    \includegraphics[width=1.0\linewidth]{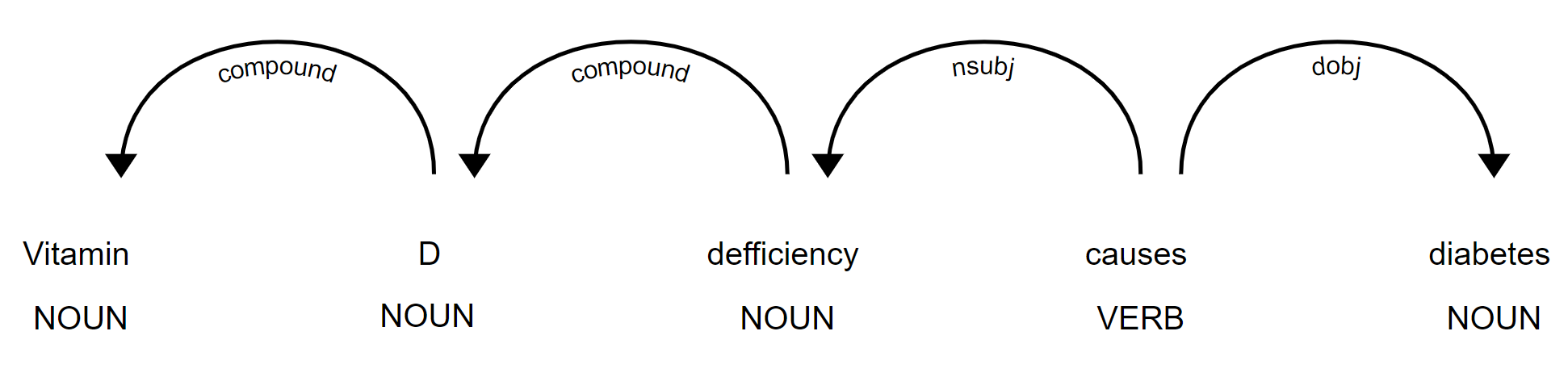}
    \caption{Dependency tree for the sentence, ``Vitamin D deficiency causes diabetes''}
    \label{fig:displacy}
\end{figure}

As shown in Figure~\ref{fig:displacy}, a dependency tree is a dependency-aware representation of a sentence. If  $\mathcal{G} = (V,E)$ is a dependency tree of $S$, then vertex-set $V$ is associated with the tokens from $S$, and 
each of the vertices is labeled with POS tag of the corresponding token. For example, for the dependency tree in Figure~\ref{fig:displacy}, the sequence of POS tags are \textit{NOUN, NOUN, NOUN, VERB,} and \textit{NOUN}. 
Edgeset $E$ represents the connecting pairs of tokens; an edge between two nodes, $i$ and $j$, denote dependency
relation between the token $s_i$ and $s_j$. \name's main motivation is to utilize the dependency information in
the transformer model. In any transformer architecture, a token receives attention from all other tokens.
Likewise, in \name, a token receives attention from other tokens in a traditional ways; besides, in a distinct
channel, a token also receives attentions from other tokens connected through dependency tree. Since the tokens
in second channel holds POS information, these POS tags are utilized in \name, allowing it to incorporate semantic information of the tokens into the model. Last, we combine the two representations of token embedding coming from two channels: traditional token-based, and dependency graph based, which is then passed to a layer for token classification. In this way, \name\ brings a flavor of graph attention network~\cite{velivckovic2017graph} in its
attention propagation mechanism to learn a better embedding of the tokens of a sentence. 

\begin{figure}[t]
    \centering
    \includegraphics[scale=0.5]{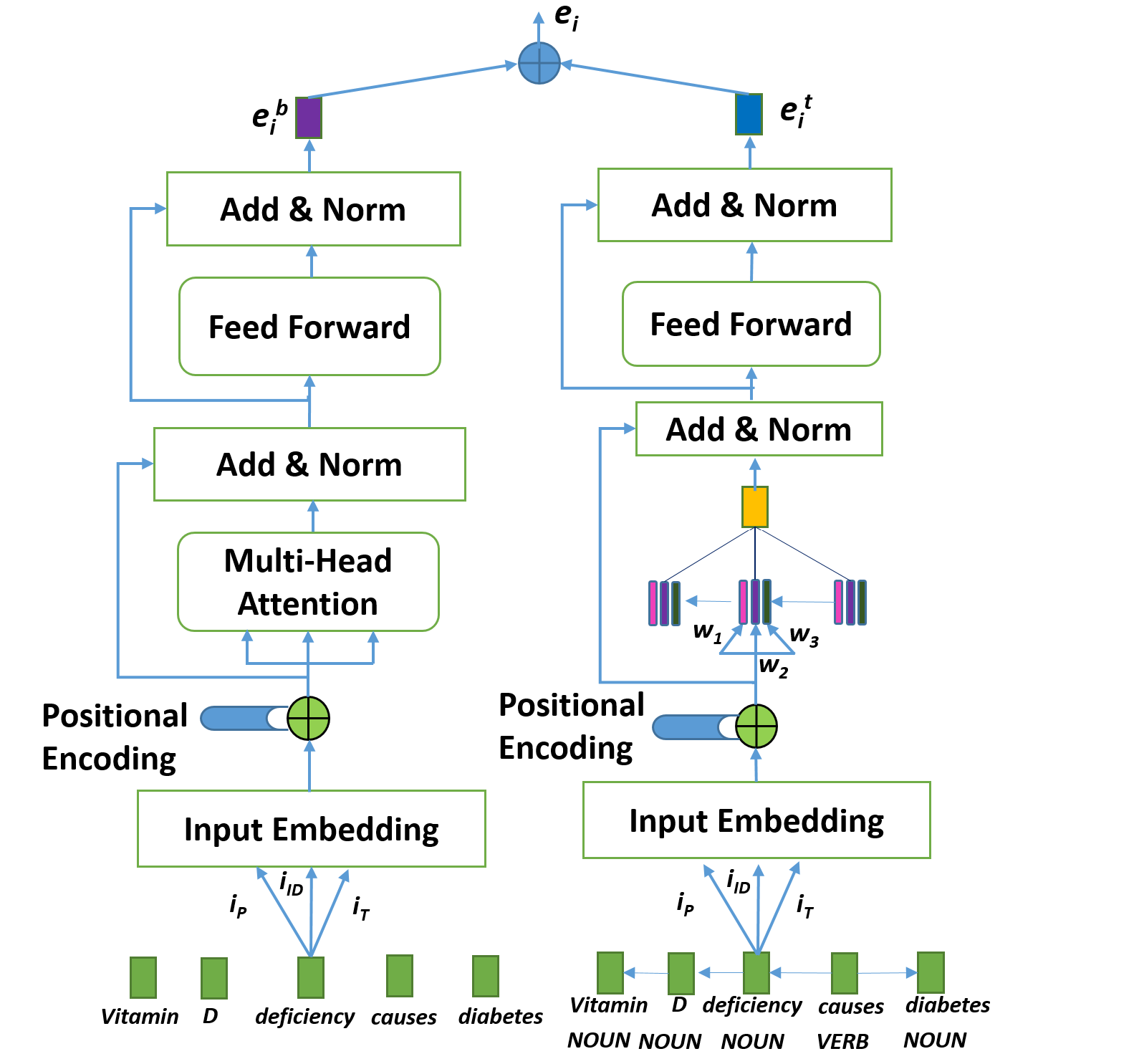}
    \caption{\name\ Model Architecture for Token Classification}
    \label{fig:classifier}
\end{figure}

\subsection{Model Architecture} \label{model}
Figure~\ref{fig:classifier} presents the token classification model of \name. The diagram showcases two main components: the token embedding encoder of a traditional BERT~\cite{bert:Embedding} on the left side and the encoder~\cite{raganato2018analysis} architecture with a modified self-attention mechanism incorporating a dependency graph on the right side. To facilitate our discussion, we utilize the running example of the text ``Vitamin D deficiency causes diabetes''. 

At the bottom of the right part of 
the architecture, we show the dependency association graph of this sentence from Spacy. As we can see the tokens for this sentence, the POS tags are: 
\{NOUN, NOUN, NOUN, VERB, and NOUN\}, from left to right. In the dependency graph there is an edge for each of the following token pairs -- \textit{(D, Vitamin), (deficiency, D), (causes, deficiency), (causes, diabetes)}. 

Our final model is a two tower model where both left and right sides learn embedding of tokens in parallel. For both towers, the standard tokenizer is modified to incorporate POS tags. To provide a more detailed illustration, \name\ generates three types of unique IDs for each token: $i_T$ for the parts of speech tag, $i_P$ for the positional embedding, and $i_{ID}$ as the input ID. These IDs are then passed to an embedding layer, and the resulting embeddings are summed to create a single vector per token, denoted as $v_i$.

For left side of the model, all representation vectors are passed through a multi-head attention layer. The multi-head attention layer learns attention for all the tokens, and the output of this layer is calculated based on the attention scores of all the tokens. This output is then normalized using a layer normalization layers and a feed forward layer to form $e_i^b$ which is the output embedding for token $s_i$ from the left part of the model.
Meanwhile, the generated tree for $S$ is fit to the right side of the diagram. Like the BERT encoder architecture, embedding of input id, and token type id, and positional id are gathered. These three embeddings are then added to construct a single vector $v_i$. The embedding of other tokens are also calculated in the similar fashion. Let the tokens connected with $s_i$ through edges is the set $\mathcal{V}$, and $s_k$ be any token in $\mathcal{V}$. Additionally, let there be three trainable matrices $\mathbf{W}_1, \mathbf{W}_2$, and $\mathbf{W}_3$. $v_i\mathbf{W}_1$, $v_i\mathbf{W}_2$, and $v_i\mathbf{W}_3$ are then query, key, and value vectors respectively for the token $s_i$. The affinity score of two connected tokens $s_i$, and $s_k$ is then calculated by the following equation.
\[a_{ik} = (v_i\mathbf{W}_1) * (v_k\mathbf{W}_2)^T\]
The attention value, $\alpha_{ik} $ (a scalar) is the softmax-score of these affinity values for all the tokens in $\mathcal{V}$ which actually represents how important the connected token is with respect to current token.

\[\alpha_{ik} = \frac{e ^ {a_{ik}}}{\sum_{j\in[1,|\mathcal{V}|]} e ^ {a_{ij}}}\]

The attention scores are used for attention based weighted sum for the output vector, $o_i$ from attention layer,
as shown in the next equation ($\mathbf{W}_4$ is another trainable matrix and $v_k\mathbf{W}_3$ is a value vector for the corresponding token.). Unlike BERT, $o_i$ is calculated for only those tokens which are connected to $s_i$. 

\[o_i = \sum_{k\in[1,|\mathcal{V}|]} \alpha_{ik} (v_k\mathbf{W}_3) \mathbf{W}_4\]

$o_i$ is then passed through a normalization layer, and the output from \textbf{Add and Norm} layer, $\bar{o_i}$ is calculated using the following equation where $\gamma$, $\beta$ are trainable scalers, $\epsilon$ is a very small scaler constant, $\mu_i$, and $\sigma^2$ are the mean, and variance of the vector $o_i$.
\[\bar{o_i} = v_i + \gamma \frac{o_i - \mu_i}{\sigma^2+\epsilon} + \beta\]
The output $\bar{o_i}$ is furthermore passed through a feed forward layer with GELU activation\cite{GELU} using another trainable matrix $\mathbf{W}_5$, and bias $b$.
\[FFN(\bar{o_i}) = GELU(\bar{o_i} \mathbf{W}_5 + b)\]
Additionally, $e_i^t$ is calculated passing the feed forward layer's output to another \textbf{Add and Norm} layer. Meanwhile, $e_i^b$ and $e_i^t$ are passed through another gate to form $e_i$ which is the token embedding of $s_i$ using
\name. This embedding is passed to another neural network for the token classification task.

\[e_i^s = \sigma(e_i^b\mathbf{W}_6 + c)\]
\[e_i = e_i^s \odot e_i^b + (1-e_i^s) \odot e_i^t\]

\section{Experiment and Result} 

We perform comprehensive experiment to show the effectiveness of \name\ for token classification task. 
Below we first discuss the dataset, and competing methods, followed by experimental results.

\subsection{Dataset} \label{dataset} 

\noindent
{\bf Semeval}:
This is a popular benchmark dataset, built by combining  the SemEval 2007 Task 4 dataset~\cite{girju-etal-2007-semeval} and the SemEval 2010 Task 8 datasets~\cite{hendrickx-etal-2010-semeval}.  A row for SemEval datasets contains a term pair, a sentence containing this pair, and the semantic relation. The SemEval 2007 Task 4 possesses 7 semantic relations whereas the SemEval 2010 Task 8 describes 9 relations. However, cause-effect relation is common among these two tasks. The datasets include predefined train and test partitions. For building validation partition, we borrow from the train partition. Train, test and validation partitions are then merged to concatenate into a single dataset. Note that, the merged dataset contains 14 relations of which we consider the instances of cause-effect relations only. The preprocessed dataset, as a result, comprises a total of 1,427 instances, all exclusively related to cause-effect relations. In terms of partition distribution, the training, test, and validation segments account for 60\%, 30\%, and 10\%, respectively.

{\bf SCITE}:
This is another dataset of the paper SCITE~\cite{li2021causality}. The dataset contains 1079 sentences exhibiting
cause-effect relation in the train dataset. We split the dataset into training, test and validation subsets maintaining  the same ratio, 6:3:1 like Semeval. Additionally, each row of this dataset also maintains the same format like the Semeval.

\noindent

\begin{table*}[h]
\centering
\caption{Example Sentences from the \dataset}
\resizebox{1.02\textwidth}{!}{%
\begin{tabular}{c|c|c}
\hline
\textbf{Cause Term} & \textbf{Effect Term} & \textbf{Sentence} \\
\hline
Diabetes & Blindness & Diabetes can lead to blindness if left uncontrolled. \\
Vitamin D deficiency & Osteoporosis & A deficiency in vitamin D can result in osteoporosis. \\
Smoking & Lung Cancer & Smoking is a major risk factor for developing lung cancer. \\
High cholesterol & Heart Disease & Elevated cholesterol levels are associated with an increased risk of heart disease. \\
Obesity & Type 2 Diabetes & Obesity is a significant risk factor for developing type 2 diabetes. \\
\hline
\end{tabular}
\label{table:example}}
\end{table*}

{\dataset}:
The previously described datasets have a very limited number of sentences as those are annotated by human. However, motivated from the recent generative models such as Bard and ChatGPT, we create a new dataset containing adequate number of sentences. To create this dataset, we harness the power of Large Language Models (LLMs) to generate cause terms, effect terms, and sentences that preserve the cause-effect relationship. It's important to note that the generated sentences may exhibit duplication and may sometimes contain special Unicode strings that require preprocessing. To ensure a wide variety of sentence structures, we develop a program capable of generating both active and passive sentence constructions. Additionally, our focus during sentence generation is on the medical domain. In total, our dataset comprises 22,273 sentences, making it a valuable resource for in-depth research into cause-effect phrase extraction. Much like SCITE, we partition this dataset randomly into training, test, and validation subsets while maintaining the same proportional distribution. In Table~\ref{table:example} we show some instances of \dataset\ dataset; The cause term, effect term, and the sentences are in Column one, two, and three respectively.

\subsection{Competing Methods} \label{rep}  For comparison, we consider a collection of neural architectures, including
BERT, Sentence-BERT, LLaMA, Dasgupta, and SCITE methods. We have also developed several models, such as BERT plus dependency and BERT plus POS tags. We discuss all of the competing methods below.

\subsubsection{Dasgupta}
Dasgupta's~\cite{dasgupta-etal-original} method is one of the first deep learning methods to extract cause-effect pairs from sentences. They design the method for token classification task. Each token is labelled either as a cause word, a effect word, causal connects or None. They learn word embedding by both word2vec
~\cite{word2vec-2013} and linguistic feature vector ~\cite{dasgupta-etal-original}. Each of the embedding is fit to a bidirectional Long Short Term Memory~\cite{Hochreiter:1997:LSTM} architecture for token classification. 

\subsubsection{SCITE}
Another method SCITE~\cite{li2021causality} uses a multi head self attention mechanism~\cite{vaswani2017attention}, and a conditional random field~\cite{fields2001probabilistic} along with a bi-LSTM architecture. Additionally, flair embedding~\cite{akbik2018contextual} is learned in a large context and transferred the string embedding for the task of causality extraction.

\begin{table*}
   \caption{Performance of \name\ compared to baseline methods in \dataset\ Dataset}  
   \setlength{\tabcolsep}{1.9pt}
        \centering
            \renewcommand{\arraystretch}{1.2}
            \scalebox{1}{
                \begin{tabular}{l | c | c| c |c}
                    \hline
                    \bf Method & \bf Prec & \bf Rec & \bf F$_1$ Score & \bf Acc (Exact)\\ 
                            &   &  & (\% imp.)  & (\% imp.) 
                    \\\hline
                    Bi-LSTM (Dasgupta) &  0.911   & 0.828  & 0.847 (-) & 0.778 (-)\\
                    Bi-LSTM-CRF (SCITE) &  0.899  & 0.834 & 0.849 (0.23) & 0.781 (0.4)\\
                    \hline
                     BERT &  0.942  & 0.958 & 0.938 (10.7)& 0.811 (4.2)\\
                     BERT plus dependency & 0.956 & 0.967 & 0.958 (13.1)& 0.833 (7.1)\\
                    BERT plus POS tags &  0.921 & 0.911 & 0.897 (5.9) & 0.831 (6.8)\\
                    \hline
                    Sentence-BERT & 0.946  & 0.964 & 0.948 (11.9) & 0.822 (5.7)\\
                    LLaMA & 0.953  & 0.956 & 0.954 (12.6) & 0.828 (6.4)\\
                      \hline
                    \textbf {\textbf{\textsc{DepBERT} (Gated)}} & \bf 0.967 & \bf 0.969 & \bf 0.963 (13.7) & \bf 0.858 (10.3)\\\hline
                \end{tabular}
            }

        \label{table:gpt}
\end{table*}

\begin{table*}
   \caption{Performance of \name\ compared to baseline methods in SemEval Dataset}  
   \setlength{\tabcolsep}{1.9pt}
        \centering
            \renewcommand{\arraystretch}{1.2}
            \scalebox{1}{
                \begin{tabular}{l | c | c| c| c}
                    \hline
        \bf Method & \bf Prec & \bf Rec & \bf F$_1$ Score & \bf Acc (Exact)\\
        
        &  &  & (\% imp.) & (\% imp.)
\\\hline
Bi-LSTM (Dasgupta) & 0.917  & 0.825  & 0.844 (-) & 0.768 (-)\\
Bi-LSTM-CRF (SCITE) & 0.896 & 0.851 & 0.86 (2) & 0.771 (0.4)\\
\hline
BERT & 0.936  & 0.941 & 0.932 (10.4) & 0.809 (5.3)\\
BERT plus dependency & 0.951 & 0.959 & 0.954 (13) & 0.841 (9.5) \\
BERT plus POS tags & 0.937  & 0.947  & 0.941 (11.5) & 0.831 (8.2) \\
\hline
Sentence-BERT & 0.926 & 0.936 & 0.933(10.5) & 0.818 (6.5)\\
LLaMA & 0.938 & 0.921 & 0.937 (11) & 0.819 (6.6)\\
\hline
\textbf{\textsc{DepBERT} (Gated)} & \bf 0.942 & \bf 0.962 & \bf 0.957 (13.4) & \bf 0.842 (9.63)\\
\hline
\end{tabular}
}
\label{table:semeval}
\end{table*}

\begin{table*}
   \caption{Performance of \name\ compared to baseline methods in SCITE Dataset}  
   \setlength{\tabcolsep}{1.9pt}
        \centering
            \renewcommand{\arraystretch}{1.2}
            \scalebox{1}{
                \begin{tabular}{l | c | c| c| c}
                    \hline
        \bf Method & \bf Prec & \bf Rec & \bf F$_1$ Score & \bf Acc (Exact) \\
        
        &  & & (\% imp.) & (\% imp.) 
\\\hline
Bi-LSTM (Dasgupta) & 0.811 & 0.825 & 0.817 (-) & 0.747 (-)\\
Bi-LSTM-CRF (SCITE) & 0.832 & 0.849 & 0.831 (1.7) & 0.751 (0.53)\\
\hline
BERT & 0.884 & 0.9 & 0.893 (9.3) & 0.768 (2.8)\\
BERT plus dependency & 0.911 & 0.923 & 0.916 (12.1) & 0.811 (8.5)\\
BERT plus POS tags & 0.908 & 0.917 & 0.906 (10.9) & 0.796 (6.5)\\
\hline
Sentence-BERT & 0.887 & 0.889 & 0.886 (8.4) & 0.773 (3.5)\\
LLaMA & 0.9 & 0.913 & 0.909 (11.3) & 0.79 (5.7)\\
\hline
\textbf{\textsc{DepBERT} (Gated)} & \bf 0.932 & \bf 0.943 & \bf 0.939 (14.9) & \bf 0.834 (11.6)\\
\hline
\end{tabular}
}
\label{table:scite}
\end{table*}

\subsubsection{BERT}
Bidirectional Encoder Representations from Transformers (BERT) is proposed by researchers from Google 
\cite{bert:Embedding}, which is not trained on any specific downstream task but instead on a more generic task called 
Masked Language Modeling. The idea is to leverage huge amounts of unlabeled data to pre-train a model,
which can be fine-tuned to solve different kinds of NLP tasks by adding a task specific layer which maps the 
contextualized token embeddings into the desired output function. In this work we use the pre-trained model ``bert-
base-uncased'' which has a vocabulary of 30K tokens and 768 dimension for each token. We use the BERT embedding for token 
classification.

\subsubsection{BERT plus dependency} We design this baseline model to incorporate the dependency structure, but unlike \name, it does not consider POS tags. In both towers of \name, we utilize the tokenizer from the original BERT model. However, this baseline model differs from BERT in that it is partially pretrained. While we do not have pretrained embeddings specifically for the dependency relation, we utilize the pretrained model ``bert-base-uncased'' for the left tower instead of training the model on an extensive corpus like Wikipedia or Google corpus.

\subsubsection{BERT plus POS Tags} We develop this baseline as well to investigate the relative importance of POS tags compared to the dependency relation. In contrast to the previous baseline, this model involves modifying the existing BERT tokenizer to include POS tokens. However, the dependency relation is not taken into account, resulting in a single tower model. Similar to the previous baseline, this model is semi-pretrained for the input ID and positional embedding.

\subsubsection{Sentence-BERT} Sentence-BERT~\cite{reimers2019sentence} is another pretrained model designed to capture mainly semantic textual similarity. The model uses siamese and triplet network structures to derive semantically meaningful sentence embedding. The model can also be used to build token representation for token classification. For 
comparison we use the pretrained weights from the model ``all-MiniLM-L6-v2'' available in Huggingface, which produces
384 dimensinoal vectors for each token.

\subsubsection{LLaMA} LLaMA~\cite{touvron2023llama} is another transformer based model developed by researchers in Meta. The model contains 7B to 65B parameters and it is trained on trillions of tokens. The model outperforms GPT 3 and other state of the art methods. The pre-trained model is available online. We use the pretrained model to represent word tokens. The representation of each tokens (4096 dimensional) is learned by LLaMA for token classification.

\subsection{Experimental Setup} Our \name\ model contains 227 millions parameters, and all of them are trainable. The left tower of the \name\ architecture is initialized with pretrained BERT parameters sourced from \textit{bert-uncased} model. It is important to note that, in our model architecture, no additional external hyperparameters are introduced. We consistently emphasize the default setup for all variations of our methods. Specifically, for LLaMA, BERT, BERT plus Dependency, BERT plus POS tags, Sentence-BERT, the number of trainable parameters stands at 524 million, 109 million, 110 million, and 110 million, respectively.  In contrast, both Dasgupta's method and SCITE feature a relatively smaller number of hyperparameters, around 400K for each. In all of our models, we make use of the Adam optimizer with a batch size of 128 and a default learning rate of 0.001. Additionally, we implement early stopping with 1000 epochs and a tolerance rate of 10, with the majority of the models concluding training within the first 100 epochs. It's worth mentioning that all the results presented in this research are derived from the initial stable runs.

\subsection{Results} 
We conducted comprehensive experiments that covered all baseline methods, including \name, across the three previously described datasets. In these experiments, accuracy was determined without allowing for partial matches. Each sentence contains a causal entity and an effect entity, and each of them may consist of one or multiple tokens. To register a correct prediction, a model needs to accurately predict all the causal and effect tokens. The results for all three datasets are conveniently presented in Tables \ref{table:gpt}, \ref{table:semeval}, and ~\ref{table:scite}. The last column in each table highlights the exact accuracy score. Additionally, our evaluation encompassed a comprehensive range of metrics, such as precision, recall, and $F_1$ score, to ensure a thorough assessment.

Table~\ref{table:gpt} presents the performance of all the methods on our \dataset\ dataset. It is observed that \name\ achieves a 10.3\% higher exact matching accuracy compared to Dasgupta's method, indicating that approximately 86\% of the extracted pairs precisely match the actual pairs. Furthermore, \name\ exhibits the best precision, recall, and $F_1$ score. Another noteworthy finding is that the BERT plus dependency method, which we designed, outperforms other baselines. This clearly demonstrates the significance of the dependency relation over POS tags. However, the combination of POS tags and the dependency relation yields even more meaningful results than solely incorporating POS tags into the model. 

Similarly, Table~\ref{table:semeval} presents the results on the SemEval dataset. The performance of all the methods on this dataset is slightly lower compared to the previous dataset. This could be attributed to the nature of cause-effect sentences. Moreover, the limited number of sentences in this dataset is insufficient to effectively train deep learning models. Nonetheless, \name\ and BERT plus dependency still outperform other methods, following a similar pattern as observed previously. In terms of $F_1$ score, \name\ achieves a 13.4\% improvement compared to Dasgupta's method.

Furthermore, Table~\ref{table:scite} displays the performance of all the baseline methods on the SCITE dataset. Due to the insufficient number of sentences in this dataset as well, the performance of all the methods falls short of expectations. However, \name\ outperforms all other baseline methods even for this dataset. The accuracy and $F_1$ score are 0.834 and 0.939, respectively, marking an improvement of 11.6\% and 14.9\% compared to Dasgupta's method.

Clearly, \name's dependency and parts-of-speech aware attention mechanism contribute to its superiority over other methods. Moreover, the combination of POS tags and the dependency relation proves to be more effective in extracting cause-effect pairs from sentences.

\section{Conclusion} 
In this study, we have effectively unveiled a pioneering method for extracting causal relationships, drawing inspiration from the sentence's underlying dependency structure. Our model, named \name, stands out by fusing the transformer architecture with dependency graph networks, harnessing the power of dependency relations and parts-of-speech markers. This amalgamation yields a marked improvement in the precision of causal relationship extraction across a multitude of domains. Looking ahead, the expansive utility of such models across various domains presents a promising path for advancing information extraction methodologies.

\section{Limitations} While the \name\ model demonstrates superior performance compared to baseline methods, it's important to note that its performance is intricately tied to the characteristics of the dataset. While large language models (LLMs) like ChatGPT can extract cause-effect pairs, even in zero-shot corpora, it's crucial to clarify that our paper does not intend to diminish the significance of LLMs. Instead, our primary emphasis lies in enhancing the foundational elements of transformer architectures to incorporate sentence dependency structures. Another limitation of our research pertains to the newly created dataset, CausalGPT. Unlike \name, which can extract multiple semantic pairs from sentences simultaneously, the CausalGPT dataset is intentionally constructed so that each sentence contains only one semantic pair. Consequently, if \name\ is trained with this dataset, it cannot extract multiple semantic pairs from sentences. That's why, in the context of this specific study, we did not conduct an evaluation of its performance in extracting multiple semantic pairs. Furthermore, we specifically emphasize semantic pairs within the English language. While semantic pairs and dependency relationships can be of great importance in all languages, it's worth noting that our research did not include experiments in languages other than English.

\section{Ethical Impacts} This research plays a pivotal role in the extraction of cause-effect relationships, negating the reliance on syntactic dependency patterns. The cause-and-effect connection serves as a foundational and indispensable element within linguistics and logic, acting as the linchpin for comprehending the intricate web of associations between events and their consequences. The extraction of causality holds paramount importance across a diverse spectrum of fields, encompassing law, medicine, and event analysis, for it provides the means to unearth the concealed mechanisms and repercussions that underlie a wide array of phenomena. Ultimately, this research empowers us to make well-informed decisions, pinpoint root causes, and enhance outcomes.

Within the legal domain, the capacity to extract cause-effect relationships without being bound by syntactic dependencies is nothing short of indispensable. The legal system hinges on precise comprehension and documentation of causality, as it is fundamental to establishing liability, attributing fault, and ensuring accountability. This research equips legal professionals with the tools to unravel the causal connections embedded in intricate legal cases, thereby simplifying the process of identifying the factors leading to specific events or circumstances. Consequently, it bolsters the pursuit of justice, whether in civil or criminal proceedings.

In the sphere of medicine, the extraction of causality fulfills a pivotal role in diagnostics and treatment. It empowers medical practitioners to discern the underlying causes of diseases and ailments, thereby facilitating more accurate and timely diagnoses. This, in turn, not only elevates the quality of patient care but also expedites the development of more effective treatment strategies. Moreover, comprehending the cause-and-effect relationships between various medical variables proves instrumental in public health endeavors, including epidemiological studies and the management of disease outbreaks.

In the context of event extraction, this research emerges as an indispensable tool across various applications, spanning disaster response, business analytics, and social science research. When grappling with extensive datasets, the identification of causality allows organizations and researchers to fathom the core drivers of specific events. In the realm of disaster response, it aids in comprehending the triggers and consequences of natural disasters, thus enhancing preparedness and response strategies. In the domain of business analytics, it facilitates the identification of factors influencing financial performance and market trends. For social science research, it provides a foundational framework for unraveling the complex dynamics that govern society, shedding light on the causes and effects of a multitude of social phenomena.

In summary, the extraction of cause-effect relationships, liberated from syntactic dependency patterns, emerges as a cornerstone in the domains of law, medicine, and event extraction, primarily due to its role in enhancing precision, accuracy, and the depth of understanding of causality within these fields. This, in turn, results in more well-informed decision-making, improved outcomes, and an increased capacity to effectively address complex challenges.

\bibliography{anthology,custom}
\bibliographystyle{acl_natbib}

\end{document}